\pdfoutput=1

\documentclass[11pt]{article}

\usepackage[preprint]{acl}

\usepackage{times}
\usepackage{latexsym}

\usepackage[T1]{fontenc}

\usepackage[utf8]{inputenc}

\usepackage{microtype}

\usepackage{inconsolata}

\usepackage{graphicx}

\usepackage{hyperref}
\usepackage{url}
\usepackage{enumitem}
\usepackage{amsmath}
\usepackage{amssymb}
\usepackage{booktabs}
\usepackage{multirow}

\usepackage{float}  

\title{Length-Controlled Margin-Based Preference Optimization without Reference Model}

\author{Gengxu Li$^{1}$ \quad Tingyu Xia$^{1}$ \quad  Yi Chang$^{1,2,3}$ \quad Yuan         
        Wu$^{1}$\footnotemark[1] \\
        $^{1}$School of Artificial Intelligence, Jilin University \\
        $^{2}$Engineering Research Center of Knowledge-Driven Human-Machine Intelligence, MOE, China \\
        $^{3}$International Center of Future Science, Jilin University\\
        gxli25@mails.jlu.edu.cn, xiaty21@mails.jlu.edu.cn, yichang@jlu.edu.cn, yuanwu@jlu.edu.cn \\
  }

\begin{document}
\maketitle

\renewcommand{\thefootnote}{\fnsymbol{footnote}}
\footnotetext[1]{Corresponding authors}


\begin{abstract}
Direct Preference Optimization (DPO) is a widely adopted offline algorithm for preference-based reinforcement learning from human feedback (RLHF), designed to improve training simplicity and stability by redefining reward functions. However, DPO is hindered by several limitations, including length bias, memory inefficiency, and probability degradation. To address these challenges, we propose Length-Controlled Margin-Based Preference Optimization (LMPO), a more efficient and robust alternative. LMPO introduces a uniform reference model as an upper bound for the DPO loss, enabling a more accurate approximation of the original optimization objective. Additionally, an average log-probability optimization strategy is employed to minimize discrepancies between training and inference phases. A key innovation of LMPO lies in its Length-Controlled Margin-Based loss function, integrated within the Bradley-Terry framework. This loss function regulates response length while simultaneously widening the margin between preferred and rejected outputs. By doing so, it mitigates probability degradation for both accepted and discarded responses, addressing a significant limitation of existing methods. We evaluate LMPO against state-of-the-art preference optimization techniques on two open-ended large language models, Mistral and LLaMA3, across ten conditional benchmarks and two open-ended benchmarks. Our experimental results demonstrate that LMPO effectively controls response length, reduces probability degradation, and outperforms existing approaches.
The code is available at \url{https://github.com/gengxuli/LMPO}.
\end{abstract}

\section{Introduction}

\begin{figure}[!t]
    \centering
    \includegraphics[width=0.8\columnwidth]{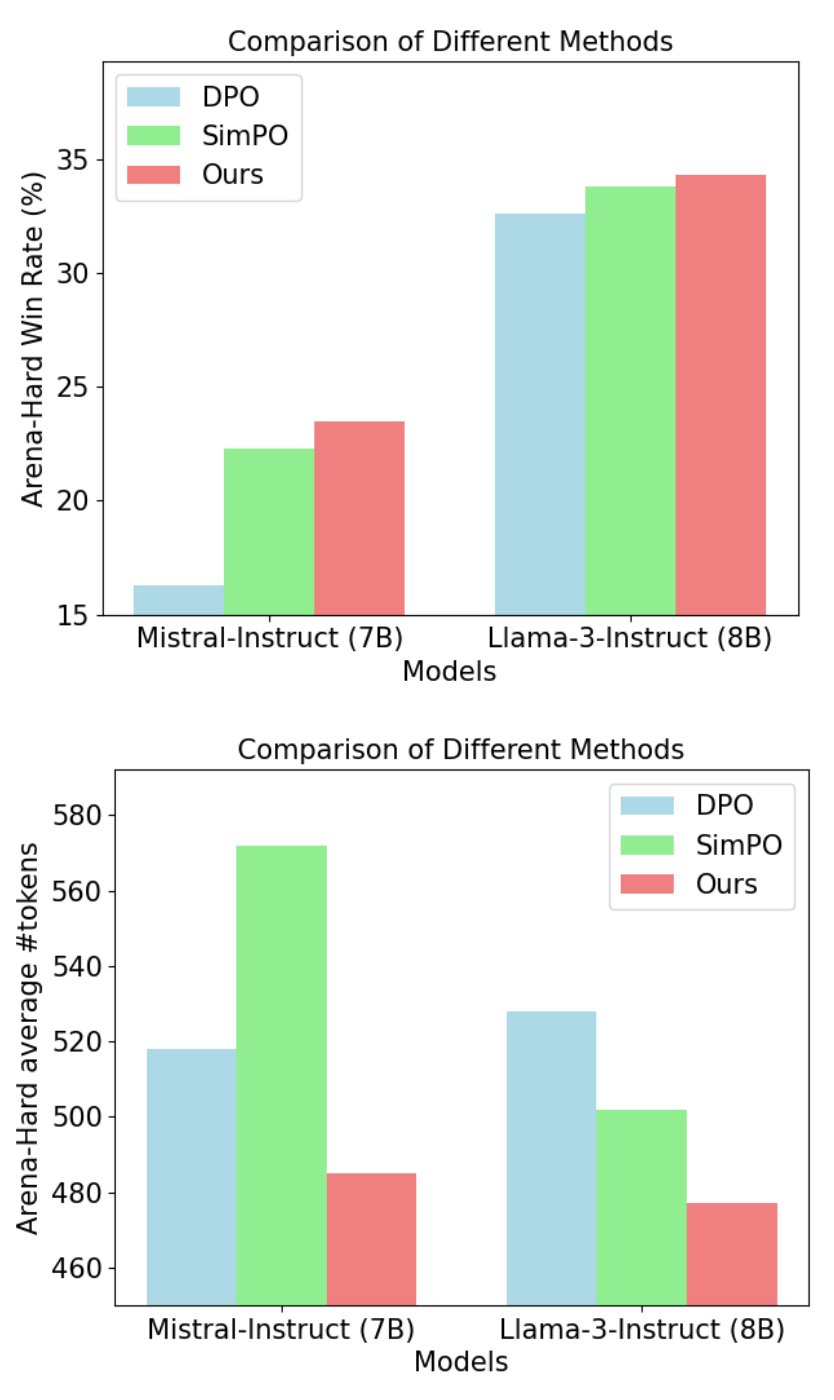}
    \caption{Comparison with DPO and SimPO under the Mistral-Instruct and Llama3-Instruct models in the Arena-Hard benchmark. Our proposed method, LMPO, achieves the highest win rate while utilizing an exceptionally low average token count across both models.}
    \label{fig:compare}
\end{figure}

Human feedback is essential for aligning large language models (LLMs) with human values and objectives~\cite{jiang2024survey,chang2024survey}, ensuring that these models act in ways that are helpful, reliable, and safe. A common strategy for achieving this alignment is reinforcement learning from human feedback (RLHF)~\cite{ziegler2019fine,stiennon2020learning,ouyang2022training}, which fine-tunes language models using human evaluations. While RLHF has shown substantial success~\cite{schulman2017proximal}, it also introduces notable challenges in optimization due to its multi-step design. This process first involves training a reward model to evaluate outputs based on human preferences, and then optimizing a policy model to maximize the assigned rewards. The complexity of these sequential steps often complicates the implementation and reduces efficiency~\cite{chaudhari2024rlhf}.

In response to these challenges, researchers have started exploring simpler alternatives that avoid the intricate, multi-stage nature of RLHF. One promising method is Direct Preference Optimization (DPO)~\cite{rafailov2024direct}, which streamlines the process by reformulating the reward function. This approach enables direct learning of a policy model from preference data, eliminating the need for a separate reward model. As a result, DPO offers greater stability and is more practical to implement.

DPO estimates implicit rewards using the log-probability ratio between a policy model’s response and that of a supervised fine-tuned (SFT) model, enabling preference learning without an explicit reward function. However, this implicit reward may misalign with the log-probability metric during inference. Moreover, DPO’s reliance on both policy and SFT models significantly increases GPU usage, especially for LLMs. The DPO loss, derived from the Bradley-Terry model, can create training imbalances, as it does not ensure an increase in the probability of positive samples—potentially reducing both positive and negative probability simultaneously. Unlike IPO ~~\cite{azar2024general}, which constrains probability variation but weakens response distinction, DPO also exhibits length bias, favoring longer responses due to preference label distribution inconsistencies~\cite{lu2024eliminating}. This issue, common in multi-stage RLHF methods, allows models to exploit verbosity for higher rewards without improving output quality, often generating responses nearly twice as long as labeled data.

To address these challenges, we introduce a novel approach incorporating a length-controlled margin-based loss function to mitigate both length bias and probability reduction. Our method consists of two key components: (1) a reference-free loss function that reduces memory inefficiency and aligns generation metrics via average log probability, and (2) a Length-Controlled Margin-Based term with two kinds of normalization methods, which minimizes probability reduction while alleviating length bias and preserving model performance. In summary, our method offers the following advantages:

\begin{itemize}[leftmargin=*]
    \item \textbf{Memory efficiency}: Our method does not rely on an extra reference model, making it more lightweight and easier to implement compared to DPO and other reference-dependent methods.
    \item \textbf{Reduction of length bias and probability decrement}: By incorporating a specially designed margin-based term, our method effectively reduces both positive and negative probability decrements, similar to traditional NLL loss, while also addressing length bias without impairing model performance.
    \item \textbf{Competitive performance}: Despite being reference-free, our method demonstrates competitive performance when compared to DPO and its variants~\cite{hong2024reference,ethayarajh2024kto}. This performance advantage is consistent across a variety of training setups and comprehensive instruction-following benchmarks, including AlpacaEval 2~\cite{li2023alpacaeval} and Arena-Hard v0.1 \cite{li2024live}.
\end{itemize}

\section{Related Work}

\paragraph{Alignment with Reinforcement Learning} 
Reinforcement learning with human feedback (RLHF) often utilizes the Bradley-Terry model~\cite{bradley1952rank} to estimate the probability of success in pairwise comparisons between two independently evaluated instances. Additionally, a reward model is trained to assign scores to these instances. Reinforcement learning algorithms, such as proximal policy optimization (PPO)~\cite{schulman2017proximal}, are used to train models to maximize the reward model's score for the selected response, ultimately enabling LLMs to align with human preferences~\cite{stiennon2020learning,ziegler2019fine}. A notable example is InstructGPT~\cite{ouyang2022training}, which showcased the scalability and adaptability of RLHF in training instruction-following language models. Alternative approaches, such as reinforcement learning with language model feedback (RLAIF~\cite{lee2023rlaif}), may also serve as feasible substitutes for human feedback~\cite{bai2022constitutional,sun2023salmon}. Nevertheless, RLHF encounters challenges, including the need for extensive hyperparameter tuning due to the instability of PPO~\cite{rafailov2024direct} and the sensitivity of the reward models~\cite{wang2024secrets}. Consequently, there is a pressing demand for more stable preference alignment algorithms.

\paragraph{Alignment Without Reward Models}
Several techniques for preference alignment reduce the reliance on reinforcement learning.
Direct Policy Optimization (DPO)~\cite{rafailov2024direct} is a method that integrates reward modeling with preference learning. And Identity Preference Optimization (IPO)~\cite{azar2024general} is introduced to mitigate potential overfitting issues in DPO.
In contrast to RLHF and DPO, an alternative approach called Kahneman-Tversky Optimization (KTO)~\cite{ethayarajh2024kto} is proposed, which does not require pairwise preference datasets.
Additionally, Preference Ranking Optimization (PRO)~\cite{song2024preference} introduces the incorporation of the softmax values from the reference response set into the negative log-probability (NLL) loss, allowing for a unified approach to supervised fine-tuning and preference alignment.

\paragraph{Alignment Without Reference Models} Due to the reliance of DPO and DPO-like methods on both the policy model and the SFT model during the alignment process, they impose greater demands on GPU resources. Several techniques have been developed to alleviate this GPU requirement by eliminating the need for a reference model. CPO~\cite{xu2024contrastive} demonstrates that the ideal loss function without a reference model can serve as the upper bound of the DPO loss, with the SFT loss acting as a replacement for the KL divergence. ORPO~\cite{hong2024reference} models the optimal reward as a log-odds function, removing the need for an additional fixed reference model. MaPO~\cite{hong2024margin} builds on the ORPO approach by introducing a margin-aware term for aligning diffusion models without a reference model. SimPO~\cite{meng2024simpo} adopts a similar reference-free preference learning framework as CPO but with improved stability due to its specific length normalization and target reward margin, leading to superior performance in various benchmarks.

\section{Method}

In this section, we begin by briefly introducing the main concept of DPO. We then propose a uniform, reference-free model based on average log-probability to address the memory and speed inefficiencies of DPO. Next, we incorporate a margin term with two kind of normalization and design a length-controlled margin-based loss function to fully leverage its benefits. Finally, we provide a detailed explanation of the margin term, illustrating how it reduces length bias and mitigates the probability decrement.

\subsection{Direct Preference Optimization (DPO)} \label{sec:three_one}

We derive our method by first revisiting Direct Preference Optimization (DPO), which offers a simplified optimization objective within the RLHF framework. DPO operates on a dataset $\mathcal{D} = { (x^{(i)}, y_w^{(i)}, y_l^{(i)}) }{i=1}^{N}$, where each input $x$ is paired with a preferred output $y_w$ and a less preferred one $y_l$. The loss function is defined as a form of maximum likelihood estimation for a policy $\pi\theta$:

\begin{equation}
\resizebox{\linewidth}{!}{%
$\begin{aligned}
\mathcal{L}(\pi_\theta;\pi_{\text{ref}}) = & -\mathbb{E}_{(x,y_w,y_l) \sim \mathcal{D}} \Big[ \log \sigma \Big( \beta \log \frac{\pi_{\theta}(y_w | x)}{\pi_{\text{ref}}(y_w | x)} \\
& \quad - \beta \log \frac{\pi_{\theta}(y_l | x)}{\pi_{\text{ref}} (y_l | x)} \Big) \Big]
\end{aligned}$%
}
\label{eq:dpo_loss}
\end{equation}

Here, $\pi_{\text{ref}}$ denotes a reference policy (typically from SFT), $\sigma$ is the sigmoid function, and $\beta$ is a scaling factor. This loss stems from a reparameterization of the reward and optimal policy, inspired by PPO. Unlike PPO, however, DPO enables training via supervised learning using only static preference-labeled data, avoiding the need for online environment interaction.

\subsection{Improvement of DPO}

\paragraph{Bradley-Terry model with home-field advantage} In Section~\ref{sec:three_one}, DPO employs the Bradley-Terry model, a well-established statistical framework frequently utilized in the analysis of competitive events, such as sporting contests. The Bradley-Terry model formalizes the human preference distribution $p^*$ as:


\begin{equation}
\resizebox{\linewidth}{!}{%
$ p^*(y_w\succ y_l \mid x)=\frac{\exp\left(r^*(x, y_w)\right)}{\exp\left(r^*(x, y_w)\right) + \exp\left(r^*(x, y_l)\right)}. $
}%
\label{eq:bradley-terry}
\end{equation}

The Bradley-Terry (BT) model utilized in DPO adopts its original formulation. However, several variants have been proposed to enhance the model’s capabilities. Notably, the Rao-Kupper model extends the BT framework by accounting for tied preferences, modeling the probability $p^*(y_w = y_l \mid x)$, which signifies that two responses, $(y_w, y_l)$, are deemed equivalent with respect to the given prompt $x$.

To better differentiate between the two responses, we reinterpret the loss response within the BT model as the "home team" in a competitive setting. Furthermore, to incorporate a potential home-field advantage, we introduce an intercept term $h$, refining the model’s capacity to capture systematic biases:



\begin{equation}
\resizebox{\linewidth}{!}{%
$\begin{aligned}
p^*(y_w\succ y_l \mid x) &= \frac{\exp\left(r^*(x, y_w)\right)}{\exp\left(r^*(x, y_w)\right) + h\exp\left(r^*(x, y_l)\right)} \\
&= \frac{1}{1 + h\exp\left(-d(x, y_w, y_l)\right)}.
\end{aligned}$%
}
\label{eq:bradley-terry-2}
\end{equation}

\paragraph{Removal of reference model} For DPO, $d(x, y_w, y_l)$ represents the term within the function $\sigma$, as outlined in Section~\ref{sec:three_one}. DPO has been widely adopted in modern models. However, despite its advantages, DPO exhibits significant drawbacks compared to standard supervised fine-tuning, such as more memory consumption and substantial computational inefficiencies due to the usage of the reference model. These limitations underscore the critical need for exploring reference model-free RLHF approaches. 

A recent approach, SimPO, utilizes the average log-likelihood function as a substitute for the reference model. However, the rationale behind this substitution remains insufficiently explained. In this work, we provide a detailed explanation to address this gap.

A recent method, CPO, demonstrates that when the reference policy $\pi_{\text{ref}}$ is set to $\pi_w$—an ideal policy that perfectly aligns with the true data distribution of preferred samples—the DPO loss $\mathcal{L}(\pi_\theta; \pi_w) + C$ is upper-bounded by $\mathcal{L}(\pi_\theta; U)$, where $C$ is a constant. Building on this result, we approximate $d(x, y_w, y_l)$ using a uniform reference model:



\begin{equation}\label{eq:loss_ref_free}
    d(x, y_w, y_l) = \log \pi_\theta(y_w|x) - \log \pi_\theta(y_l|x).
\end{equation}

Next, in DPO and CPO, the implicit reward is defined as the log ratio of the response probability between the current policy model and the SFT model. However, this reward formulation does not directly align with the metric guiding generation, which is roughly the average log probability of a response generated by the policy model. This discrepancy between the training and inference phases may negatively impact performance. To address this, we replace the log probability with the average log probability in Eq.~\ref{eq:loss_ref_free}:


\begin{equation}
\resizebox{\linewidth}{!}{%
$ d(x, y_w, y_l) = \frac{\beta}{|y_w|} \log \pi_\theta(y_w|x) - \frac{\beta}{|y_l|} \log \pi_\theta(y_l|x). $
}%
\label{eq:loss_avg}
\end{equation}

\subsection{Length-Controlled Margin-Based Loss}

To ensure a more pronounced separation in reward scores for responses with greater quality differences, we incorporate a margin term into the Bradley-Terry framework. The modified objective is as follows:

\begin{equation}
\resizebox{\linewidth}{!}{%
$ d(x, y_w, y_l) = r^*(x, y_w) - r^*(x, y_l) - \lambda m(y_w,y_l,x). $
}%
\label{equ:cpmod}
\end{equation}

Here, \( m(y_w, y_l, x) \) denotes the margin quantifying the preference strength between the winning response \( y_w \) and the losing response \( y_l \) given input \( x \), and \(\lambda\) is a scaling factor. The function \( r^*(x, y) \) provides the reward score for response \( y \) conditioned on input \( x \). Incorporating this margin enables the model to better differentiate reward scores, especially when the quality gap between responses is large.

Recent works have adopted this formulation to improve performance. For instance, the reward models in Llama-2-Chat \citep{touvron2023llama} and UltraRM \citep{cui2023ultrafeedback} use discrete preference scores as margin terms, while SimPO \citep{meng2024simpo} employs a fixed margin to ensure the preferred response always receives a higher reward than the less favored one. Nevertheless, issues such as length bias remain.

To address these challenges, we propose the Length-Controlled Margin-Based Loss. This loss explicitly regulates the length of generated responses, mitigating the tendency of large language models to prefer longer outputs. It also controls the probability decrease for both selected and rejected responses, enhancing the model’s ability to distinguish between correct and incorrect answers. Importantly, it enlarges the margin between the probabilities of chosen and rejected responses, strengthening the model’s discrimination of response quality. The full formulation is given below.

\begin{equation}
\resizebox{\linewidth}{!}{%
$ m(x,y_w,y_l) = (1 - p_\theta(y_w|x)) \cdot \left(1 - (p_\theta(y_w|x) -  p_\theta(y_l|x))^5 \right). $
}%
\label{eq:loss_margin}
\end{equation}

In neural machine translation (NMT) adequacy evaluation, the use of the "power of 5" in the margin term has been shown to be the most effective approach. Prior studies~\cite{miao2021prevent} have demonstrated its superiority through ablation experiments comparing various margin formulations. Additionally, the "power of 5" margin has been incorporated into recent Mixture-of-Experts (MoE) models, such as MoE-Summ~\cite{chen2024flexible}, achieving significant improvements across multiple tasks. Motivated by these findings, we adopt the "power of 5" margin term in this work.

\noindent \textbf{Normalization}: To enhance training stability and regulate the length of model outputs, we employ two distinct normalization techniques: average length normalization and Z-score normalization \citep{patro2015normalization}.

(1) average length normalization: To mitigate length bias in LLM-generated outputs, we introduce a dynamic scaling factor, defined as $\frac{|y_w| + |y_l|}{2 * |y|}$ to adjust the rewards for both chosen and rejected outputs. This factor is incorporated into Eq.~\ref{eq:loss_margin}, modifying the probability formulation as follows:

\begin{equation}
    p_\theta(y|x) = \exp\left( \frac{1}{|y|}\log \pi_\theta(y|x) * \frac{|y_w| + |y_l|}{2 * |y|}\right)
\end{equation}

(2) Z-score normalization: To stabilize training and prevent the loss from being dominated by scale variations in $m(y_w,y_l,x)$, we apply Z-score normalization to $m$, yielding:

\begin{equation}
    \overline{m}(x, y_w, y_l) = \frac{m(x, y_w, y_l) - a_m}{b_m},
\end{equation}
where $a_m$ and $b_m$  denote the mean and standard deviation of $m$ computed over the entire training process.

\noindent \textbf{Objective.} Finally, we obtain the LMPO final loss function by incorporating the above considerations: 

\begin{equation}
\resizebox{\linewidth}{!}{%
$ \mathcal{L}_{\text{LMPO}}(\pi_\theta) = - \mathbb{E}_{(x, y_w, y_l) \sim \mathcal{D}}\left[ \log \left( \frac{1}{1 + h\exp\left(-d(x, y_w, y_l)\right)} \right) \right]. $
}%
\label{eq:lmpo}
\end{equation}

where 

\begin{equation}
\resizebox{\linewidth}{!}{%
$\begin{aligned}
d(x, y_w, y_l) &= \frac{\beta}{|y_w|} \log \pi_\theta(y_w|x) - \frac{\beta}{|y_l|} \log \pi_\theta(y_l|x) \\
&\quad - \lambda \overline{m}(x, y_w, y_l).
\end{aligned}$%
}
\label{eq:lmpo_with_m}
\end{equation}

In summary, LMPO employs an implicit reward formulation that directly aligns with the generation metric, eliminating the need for a reference model. Next, it introduces a margin term $\text{$m(\mathbf{x},\mathbf{y}^w,\mathbf{y}^l)$}$ with two kinds of normalization methods to help separate the winning and losing responses, alleviate length bias and wining response probability decrement problems.   Details of  Z-score normalization and further analysis of LMPO loss are shown in Appendix~\ref{sec:gradient}.

\section{Experiment}

\subsection{Experimental Setup}
\textbf{Models and Training Settings.}  
We optimize preferences using two model families: Llama3-8B \citep{llama3modelcard} and Mistral-7B \citep{jiang2023mistral}, under two setups: Base and Instruct.

In the Base setup, following SimPO, we use pre-trained SFT models \href{https://huggingface.co/alignment-handbook/zephyr-7b-sft-full}{Zephyr-7B-SFT}~\citep{tunstall2023zephyr} and \href{https://huggingface.co/princeton-nlp/Llama-3-Base-8B-SFT}{Llama-3-Base-8B-SFT} as initialization. Preference optimization is then performed on the UltraFeedback dataset \citep{cui2023ultrafeedback}, which contains feedback from LLMs of varying quality.

For the Instruct setup, we use instruction-tuned models \href{https://huggingface.co/mistralai/Mistral-7B-Instruct-v0.2}{Mistral-7B-Instruct-v0.2} and \href{https://huggingface.co/meta-llama/Meta-Llama-3-8B-Instruct}{Meta-Llama-3-8B-Instruct} as SFT models. We adopt the same training data as SimPO: \href{https://huggingface.co/datasets/princeton-nlp/mistral-instruct-ultrafeedback}{princeton-nlp/llama3-ultrafeedback} and \href{https://huggingface.co/alignment-handbook/zephyr-7b-sft-full}{princeton-nlp/mistral-instruct-ultrafeedback} for Llama3-8B and Mistral-7B, respectively.

These settings reflect recent advances, placing our models among top performers on several leaderboards.

\begin{table*}[!t]
\centering
\small 
\caption{AlpacaEval 2 and Arena-Hard results under the four settings. LC and WR denote length-controlled and raw win rate, respectively. Length denotes the length of the generated prompt. We train SFT models for Base settings on the UltraChat dataset. For Instruct settings, we follow the training process of SimPO.}

\resizebox{0.9\textwidth}{!}{
\begin{tabular}{lcccccccccc}
\toprule
\multirow{3}{*}{\textbf{Method}} & \multicolumn{5}{c}{\textbf{Mistral-Base (7B)}} & \multicolumn{5}{c}{\textbf{Mistral-Instruct (7B)}} \\ 
\cmidrule(lr){2-6}\cmidrule(lr){7-11}
& \multicolumn{3}{c}{\textbf{AlpacaEval 2}} & \multicolumn{2}{c}{\textbf{Arena-Hard}}  & \multicolumn{3}{c}{\textbf{AlpacaEval 2}} & \multicolumn{1}{c}{\textbf{Arena-Hard}} \\
\cmidrule(lr){2-4}\cmidrule(lr){5-6} \cmidrule(lr){7-9}\cmidrule(lr){10-11} 
& {\scriptsize \bf LC (\%)} & {\scriptsize \bf WR (\%)} & {\scriptsize \bf Length} & {\scriptsize \bf WR (\%)} & {\scriptsize \bf Length}  & {\scriptsize \bf LC (\%)}  & {\scriptsize \bf WR (\%)} & {\scriptsize \bf Length}  & {\scriptsize \bf WR (\%)} & {\scriptsize \bf Length}  \\
\midrule
SFT       & 6.2   & 4.6   & 1082   & 3.3  & 437   & 17.1   & 14.7   & 1676   & 12.6   & 486     \\
\midrule
DPO       & 15.1  & 12.5  & 1477  & 10.4  & 628   & 26.8   & 24.9   & 1808   & 16.3   & 518     \\
SLiC      & 10.9  & 8.9   & 1525  & 7.3   & 683   & 24.1   & 24.6   & 2088   & 18.1   & 517     \\
IPO       & 11.8  & 9.4   & 1380  & 7.5   & 674   & 20.3   & 20.3   & 2024   & 16.2   & 740     \\
CPO       & 9.8   & 8.9   & 1827  & 5.8   & 823   & 23.8   & 28.8   & 3245   & 22.6   & 812     \\
KTO       & 13.1  & 9.1   & 1144  & 5.6   & 475   & 24.5   & 23.6   & 1901   & 17.9   & 496      \\
SimPO     & 17.7  & 16.5  & 1803  & 14.3  & 709   & 29.7   & 31.7   & 2350   & 22.3   & 572      \\ 
\midrule
LMPO      & 20.9  & 14.9  & 1351  & 13.8  & 458  & 29.8    & 28.0   & 1881   & 23.5   & 485      \\
\midrule[.7pt]
\multirow{3}{*}{\textbf{Method}} & \multicolumn{5}{c}{\textbf{Llama-3-Base (8B)}} & \multicolumn{5}{c}{\textbf{Llama-3-Instruct (8B)}} \\ 
\cmidrule(lr){2-6}\cmidrule(lr){7-11}
& \multicolumn{3}{c}{\textbf{AlpacaEval 2}} & \multicolumn{2}{c}{\textbf{Arena-Hard}}  & \multicolumn{3}{c}{\textbf{AlpacaEval 2}} & \multicolumn{1}{c}{\textbf{Arena-Hard}} \\
\cmidrule(lr){2-4}\cmidrule(lr){5-6} \cmidrule(lr){7-9}\cmidrule(lr){10-11} 
& {\scriptsize \bf LC (\%)} & {\scriptsize \bf WR (\%)} & {\scriptsize \bf Length} & {\scriptsize \bf WR (\%)} & {\scriptsize \bf Length}  & {\scriptsize \bf LC (\%)}  & {\scriptsize \bf WR (\%)} & {\scriptsize \bf Length}  & {\scriptsize \bf WR (\%)} & {\scriptsize \bf Length}  \\
\midrule
SFT     &  8.4  &  6.2  &  914  &  1.3  &  521  & 26.0  & 25.3  & 1920  & 22.3  &  596  \\
\midrule
DPO     & 18.2  & 15.5  & 1585  & 15.9  &  563  & 40.3  & 37.9  & 1883  & 32.6  &  528  \\
SLiC    & 12.1  & 10.1  & 1540  & 10.3  & 676   & 31.3  & 28.4  & 1805  & 26.5  & 502 \\
IPO     & 14.4  & 14.2  & 1856  & 17.8  &  608  & 35.6  & 35.6  & 1983  & 30.5  &  554  \\
CPO     & 12.3  & 13.7  & 2495  & 11.6  &  800  & 28.9  & 32.2  & 2166  & 28.8  &  624  \\
KTO     & 14.2  & 12.4  & 1646  & 12.5  &  519  & 33.1  & 31.8  & 1909  & 26.4  &  536  \\
SimPO   & 21.6  & 20.0  & 1818  & 26.9  &  877  & 43.9  & 39.0  & 1788  & 33.8  &  502  \\ 
\midrule
LMPO    & 21.3  & 17.7  & 1601  & 30.1  & 1114  & 43.7  & 39.0  & 1791  & 34.3  &  477  \\  
\bottomrule
\end{tabular}
}
\label{tab:main_res}
\vspace{-1.5em}
\end{table*}

\noindent \textbf{Evaluation Benchmarks.} We evaluate our models using two widely recognized open-ended instruction-following benchmarks: AlpacaEval 2 \citep{li2023alpacaeval} and Arena-Hard v0.1 \citep{li2024live}. These benchmarks evaluate the models' conversational abilities across a wide range of queries and are widely used by the research community \citep{chang2024survey}. For AlpacaEval 2, we report both the raw win rate (WR) and the length-controlled win rate (LC) \citep{dubois2024length}, with the LC metric designed to mitigate the effects of model verbosity. For Arena-Hard, we report the win rate (WR) against a baseline model. 

Additionally, we evaluate the models on ten downstream tasks in the Huggingface Open Leaderboard V1 and V2, following SimPO \citep{meng2024simpo} and SIMPER~\cite{xiao2025simper}. These downstream tasks include the AI2 Reasoning Challenge (25-shot) \citep{clark2018think}, TruthfulQA (0-shot) \citep{lin2021truthfulqa}, Winogrande (5-shot) \citep{sakaguchi2021winogrande}, GSM8K (5-shot) \citep{cobbe2021training}, IFEval~\cite{zhou2023instruction}, BBH~\cite{suzgun2022challenging}, MATH~\cite{hendrycks2021measuring}, GPQA~\cite{rein2024gpqa}, MuSR~\cite{sprague2023musr}, MMLU-PRO~\cite{wang2024mmlu}. We report the match accuracy for these conditional benchmarks. Additional details are provided in Appendix \ref{Appendix:A}.

\noindent \textbf{Baselines} We perform a comparative analysis of our method against several state-of-the-art offline preference optimization techniques, including DPO \citep{rafailov2024direct}, SLiC~\cite{zhao2023slic}, IPO \citep{azar2024general}, CPO \citep{xu2024contrastive}, KTO \citep{ethayarajh2024kto} and SimPO \citep{meng2024simpo}. For SimPO, we use the model provided for the Llama3-8B family and replicate the SimPO methodology for the Mistral-7B family in our environment. For the other methods, we report the results provided by SimPO. We also tune the hyperparameters for SimPO and report the best performance achieved.

\subsection{Main Results}

\textbf{LMPO achieves a favorable trade-off between performance and prompt efficiency across multiple benchmarks.} As shown in Table~\ref{tab:main_res}, while all preference optimization methods improve upon the SFT baseline, LMPO demonstrates competitive results, particularly on AlpacaEval 2 and Arena-Hard, with a clear advantage in controlling prompt length.

\textbf{AlpacaEval 2.} LMPO generates significantly shorter prompts than SimPO in three evaluated settings. For example, in the Mistral-Base (7B) setting, LMPO outperforms SimPO by 3.2\% on the LC metric despite using much shorter prompts. Although LMPO may not achieve the highest scores on LC and WR, its ability to maintain competitive performance with shorter outputs highlights its efficiency. This indicates that LMPO achieves a meaningful trade-off between performance and prompt length, making it a practical option for scenarios where both generation quality and inference speed are important.

\textbf{Arena-Hard.} LMPO obtains the highest win rate among all compared methods, while still maintaining shorter prompt lengths. This showcases its effectiveness in more challenging settings, where both accuracy and efficiency are critical. Interestingly, in the Llama-3-Base (8B) configuration, LMPO’s prompt length is noticeably longer. This is likely caused by tokenizer-related artifacts (e.g., the presence of multiple BOS tokens), which can affect the computed length but not the model’s core efficiency.

\textbf{Overall.} LMPO achieves strong performance on both AlpacaEval 2 and Arena-Hard, with particularly notable results on the latter benchmark. The difference in improvements across the two datasets may stem from their distinct characteristics—Arena-Hard contains more complex and diverse tasks than AlpacaEval 2. LMPO’s stronger results on Arena-Hard further confirm its suitability for handling difficult problems, demonstrating its advantage in complex real-world scenarios. These results suggest that LMPO is a practical and effective approach that balances concise outputs with solid performance across diverse evaluation settings.

\begin{table}[!t]
\centering
\small 
\caption{Ablation studies under Llama-3-Base (8B) settings. We report the win rate and 95\% confidence interval for Arena-Hard.}
\vspace{-0.4em}
\label{tab:ablation}
\resizebox{0.5\textwidth}{!}{
\begin{tabular}{l*{10}{c}}
\toprule
\multirow{2}{*}{\textbf{Method}}
& \multicolumn{4}{c}{\textbf{Arena-Hard}} \\
\cmidrule(lr){2-5}
& {\scriptsize \bf WR (\%)} & {\scriptsize \bf 95 CI high (\%)} & {\scriptsize \bf 95 CI low  (\%)} 
& {\scriptsize \bf Length} \\
\midrule
SimPO  & 26.9        & 28.7               & 25.1   & 877 \\
\midrule
LMPO & 30.1        & 32.4               & 27.7  & 1114\\
\midrule
w/o Z-score normalization  & 22.5        & 25.0                & 20.0             & 630 \\
w/o avg-length normalization  & 27.9         & 29.6                & 26.2             & 843 \\
log function & 27.9         & 30.1               & 25.9              & 770 \\
cube function & 29.3        & 31.7                & 27.4              & 903 \\
sigmoid function & 25.2         & 27.3                & 22.5             & 649\\

\bottomrule
\end{tabular}
}
\vspace{-1em}
\end{table}

\noindent \textbf{The importance of the design on the loss term.}
As the core contribution of LMPO is to propose a novel loss term $ m(x,y_w,y_l) = (1 - p_\theta(y_w|x)) \cdot \left(1 - (p_\theta(y_w|x) -  p_\theta(y_l|x))^5 \right)$, we also evaluate other variants of the reference model. Specifically, we compare LMPO with three variants: 

\begin{itemize}[leftmargin=*]
    \item log function: $ m(x,y_w,y_l) = (1 - p_\theta(y_w|x)) \cdot \left(\frac{1}{\alpha}log(\frac{1-(p_\theta(y_w|x) -  p_\theta(y_l|x))}{1+(p_\theta(y_w|x) -  p_\theta(y_l|x))})+0.5\right)$
    \item cube function: $ m(x,y_w,y_l) = (1 - p_\theta(y_w|x)) \cdot \left(1 - (p_\theta(y_w|x) -  p_\theta(y_l|x))^3 \right)$
    \item sigmoid function: $ m(x,y_w,y_l) = (1 - p_\theta(y_w|x)) \cdot \left(\frac{1}{1 + \exp({\frac{p_\theta(y_w|x) -  p_\theta(y_l|x)}{\beta}})}\right)$
\end{itemize}
where $\alpha$ is a hyperparamater for log function and $\beta$ is a hyperparamater for sigmoid function.

As shown in Table~\ref{tab:ablation}, most of the variants outperform SimPO, highlighting the significance of the loss term. Furthermore, our proposed reference model consistently exceeds the performance of other variants, demonstrating the effectiveness of the proposed design. However, the prompt length of our loss term is the longest among the options, which may affect performance. The log function achieves better performance with a shorter length compared to SimPO. Therefore, exploring improved loss functions will be a key direction for future experiments in LMPO.

\noindent \textbf{All key designs in LMPO are crucial.}
To further assess the impact of various components in LMPO, we conduct ablation studies by removing key elements. As shown in Table~\ref{tab:ablation}, removing Z-score normalization and average-length normalization leads to significant performance drops, underscoring the importance of these components in LMPO. However, removing these two terms reduces the prompt length, suggesting a need to balance model performance with prompt length. Additionally, due to resource limitations, certain aspects of LMPO, such as the home-court advantage, were not removed, which presents an opportunity for future research.

\section{Analysis}

\subsection{Reduction of probability decrement}

First we introduce the loss function SimPO, the loss function for SimPO is formulated as a maximum likelihood estimation for a policy model parameterized by  $\pi_\theta$:

\begin{equation}
\resizebox{\linewidth}{!}{%
$\begin{aligned} 
\mathcal{L}_{\textbf{SimPO}}(\pi_\theta) = & - \mathbb{E}_{(x, y_w, y_l) \sim \mathcal{D}}\left[ \log \sigma \left( \frac{\beta}{|y_w|} \log \pi_\theta(y_w|x) \right. \right. \\ & \left. \left. - \frac{\beta}{|y_l|} \log \pi_\theta(y_l|x) - \gamma \right) \right]. 
\end{aligned}$%
}
\label{eq:dpo_loss_new}
\end{equation}

where $\gamma$ is a hyperparameter call target reward margin, which is a constant with no gradient.

The primary optimization objective in Eq. \ref{eq:dpo_loss_new} is to maximize the margin between the chosen and rejected probabilities, without directly controlling either of them. This lack of control may result in a reduction in both probabilities during training. Furthermore, a decrease in the chosen probability contradicts the goal of aligning the language model with human preferences.

\begin{figure}[t]
  \resizebox{\columnwidth}{!}{\includegraphics{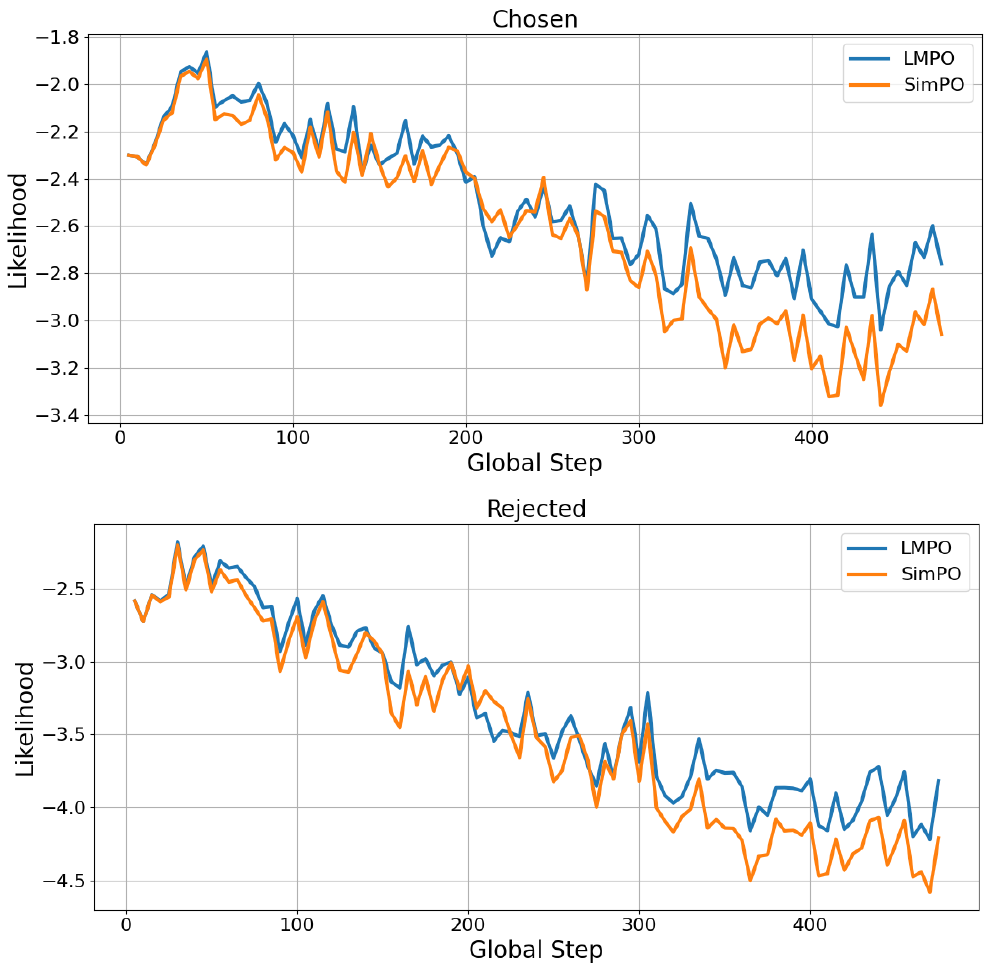}}
  \caption{The curves of the chosen (top) and rejected (bottom) log-probabilities during the training process in the Llama-3-Base (8B) setting. The \textcolor{blue}{blue} and \textcolor{orange}{orange} curves represent LMPO and SimPO, respectively.}
  \label{fig:constraint}
\end{figure}

In LMPO, we introduce a constraint term, $1 - p_\theta(y_w|x)$. By minimizing the loss function, LMPO effectively maximizes the exponentiated log-probability, implicitly imposing a constraint on the log-probability. It is worth noting that the constraint term we use is similar to the SFT loss employed in CPO \citep{xu2024contrastive}. However, relying solely on the SFT loss can impose an excessive constraint, which may negatively impact the performance of the method. Therefore, we combine the latent constraint term with a margin term to balance the reduction of probability decrement while maximizing the margin.

As shown in Figure \ref{fig:constraint}, it is evident that LMPO imposes a constraint on the log-probabilities of both chosen and rejected responses, in contrast to SimPO. Despite this constraint, LMPO is still able to maximize the margin between these two probabilities, with the margins being similar to those of SimPO. By reducing the probability decrement and maximizing the margin, LMPO can achieve competitive performance when compared to SimPO.

\subsection{Hyperparameter Selection}

As shown in Eq. \ref{eq:lmpo_with_m}, LMPO employs a hyperparameter $\lambda$ to control the margin loss term. Additionally, since Z-score normalization is applied to compute the overall margin loss during the training process, adjusting $\lambda$ can significantly affect $\overline{m}(x, y_w, y_l)$, thereby influencing the model’s preferences.

We selected three values for the hyperparameter $\lambda$: 0.05, 0.2, and 1.0, and applied them to the LMPO algorithm under the Mistral-Base (7B) setting. The results of AlpacaEval 2 are presented in Table \ref{tab:hyperparameter}. It is evident that as $\lambda$ increases, the WR remains relatively stable, while the LC increases with $\lambda$, and the length of the generated prompt decreases. These findings suggest that LMPO has a notable impact on prompt length control and performs well in scenarios requiring length regulation.

\begin{table}[!t]
\centering
\small 
\caption{AlpacaEval 2 results for Hyperparameter Selection under Mistral-Base (7B) settings. LC and WR denote length-controlled and raw win rate, Length denotes the length of the generated prompt, STD means standard deviation of win rate.}
\vspace{-0.4em}
\label{tab:hyperparameter}
\resizebox{0.5\textwidth}{!}{
\begin{tabular}{l*{10}{c}}
\toprule
\multirow{2}{*}{\textbf{Method}}
& \multicolumn{4}{c}{\textbf{AlpacaEval 2}} \\
\cmidrule(lr){2-5}
& {\scriptsize \bf Lc (\%)} & {\scriptsize \bf WR (\%)} & {\scriptsize \bf STD  (\%)} 
& {\scriptsize \bf Length} \\

\midrule

$\lambda$=0.05 & 16.1         & 14.6               & 1.1             & 1751 \\
$\lambda$=0.2 & 16.6        & 15.0                & 1.0              & 1726 \\
$\lambda$=1.0 & 20.9         & 14.9                & 1.1             & 1351\\

\bottomrule
\end{tabular}
}
\vspace{-1em}
\end{table}

\begin{figure}[t]
  \resizebox{\columnwidth}{!}{\includegraphics{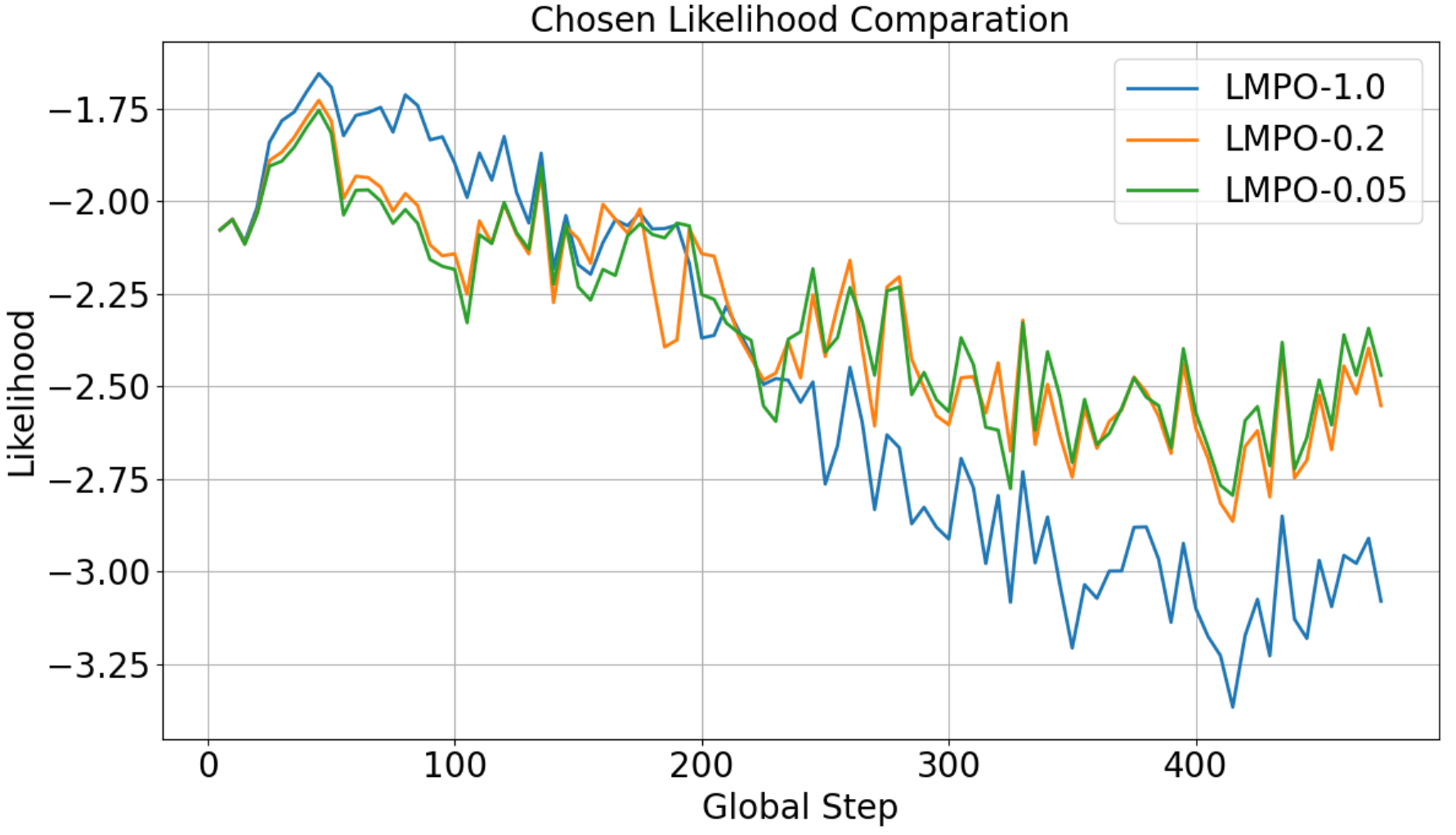}}
  \caption{The curves of the chosen log-probabilities during the training process in the  Mistral-Base (7B) setting. The red,  green and blue curves represent $\lambda$=0.05, $\lambda$=0.2 and $\lambda$=1.0, respectively.}
  \label{fig:hyperparameter}
\end{figure}

To demonstrate the effect of hyperparameter selection on the reduction of probability decrement, we present the training curves for these three training processes. The results are shown in Figure \ref{fig:hyperparameter}. It is clear that as $\lambda$ increases, the log-probabilities of the selected prompts decrease significantly, and the corresponding curves decline rapidly. These findings indicate that increasing $\lambda$ may adversely affect the latent constraint mechanism in LMPO, which is undesirable for its intended performance.

Therefore, selecting an appropriate hyperparameter for LMPO is crucial, as it depends on the specific scenario. Choosing an optimal hyperparameter can strike a balance between achieving better performance in a length-controlled setting and minimizing the reduction in probability decrement.

\section{Conclusion}

In this paper, we introduce LMPO, which uses a length-controlled margin-based loss function to mitigate length bias and probability reduction. It features a reference-free loss for memory efficiency and a margin-based term with two normalization methods to balance probability control and model performance. Without requiring a reference model, it remains lightweight while effectively reducing length bias and probability decrement. Despite its simplicity, the method achieves competitive results compared to DPO and its variants across multiple benchmarks, including two open-ended benchmarks: AlpacaEval 2, Arena-Hard v0.1 and ten conditional benchmarks used in Huggingface open leaderboard V1 and V2. 

\section*{Limitations} 

The constraints of LMPO are outlined as follows: 

\paragraph{Settings.}The settings we use in our paper are based on those from the early version of SimPO. In later versions, SimPO adopts other configurations, such as Llama-3-Instruct v0.2 and Gemma. For a more in-depth analysis, updating the settings is necessary.

\paragraph{Performance.} LMPO does not outperform SimPO in AlpacaEval 2 and struggles with downstream tasks, particularly underperforming in mathematical settings like GSM8K. To improve its performance, further updates are needed, such as selecting a better loss function and employing more effective normalization methods. Additionally, the updated Llama3 tokenizer occasionally introduces two BOS tokens, which can impact evaluation results. For example, this causes an unusually long generated prompt for LMPO in AlpacaEval 2 under the Llama-3-Base setting. Therefore, it may be necessary to use the pre-update Llama3 tokenizer.


\bibliography{custom}

\appendix

\section{Comprehensive Gradient Analysis and Justification of LMPO} \label{sec:gradient}

We provide a detailed gradient derivation of the LMPO loss to clarify how it improves separation between winning and losing responses, mitigates length bias, and preserves winning response probabilities during training.

\paragraph{1. LMPO Loss Recap.}

The LMPO loss is defined as

\begin{equation}
\resizebox{\linewidth}{!}{$
\mathcal{L}_{\text{LMPO}} = 
-\mathbb{E}_{(x,y_w,y_l) \sim \mathcal{D}} 
\left[ \log \left( \frac{1}{1 + h \cdot \exp\bigl(-d(x,y_w,y_l)\bigr)} \right) \right]
$}
\end{equation}

where

\begin{equation}
\begin{aligned}
d(x,y_w,y_l) =\ & \frac{\beta}{|y_w|} \log \pi_\theta(y_w|x) 
- \frac{\beta}{|y_l|} \log \pi_\theta(y_l|x) \\
& - \lambda\, \overline{m}(x,y_w,y_l)
\end{aligned}
\end{equation}

\paragraph{2. Gradient of the Loss w.r.t. $d$.}

Denote the sigmoid function as $\sigma(z) = \frac{1}{1+e^{-z}}$. Then,

\begin{equation}
\begin{aligned}
\frac{\partial \mathcal{L}_{\text{LMPO}}}{\partial d}
= {} & \frac{\partial}{\partial d} \log(1 + h \cdot e^{-d}) \\
= {} & - \frac{h \cdot \exp(-d)}{1 + h \cdot \exp(-d)} \\
= {} & \sigma(-d + \log h) - 1
\end{aligned}
\end{equation}

For simplicity, assuming $h=1$,
\begin{equation}
\frac{\partial \mathcal{L}_{\text{LMPO}}}{\partial d} = - \frac{\exp(-d)}{1 + \exp(-d)} = \sigma(-d) - 1.
\end{equation}

\paragraph{3. Gradient of $d$ w.r.t. Model Parameters $\theta$.}

Since $d$ depends on $\pi_\theta$ through the log-probabilities and the margin term, its gradient is:

\begin{equation}
\begin{aligned}
\nabla_\theta d =\ & \frac{\beta}{|y_w|} \nabla_\theta \log \pi_\theta(y_w|x) 
- \frac{\beta}{|y_l|} \nabla_\theta \log \pi_\theta(y_l|x) \\
& - \lambda\, \nabla_\theta \overline{m}(x,y_w,y_l)
\end{aligned}
\end{equation}

\paragraph{4. Gradient of Log-Probability Terms.}

Recall that
\begin{equation}
\log \pi_\theta(y|x) = \sum_{t=1}^{|y|} \log \pi_\theta(y_t | y_{<t}, x).
\end{equation}

Then,
\begin{equation}
\nabla_\theta \log \pi_\theta(y|x) = \sum_{t=1}^{|y|} \nabla_\theta \log \pi_\theta(y_t | y_{<t}, x).
\end{equation}

Dividing by length $|y|$ normalizes this gradient by sequence length, mitigating length bias by ensuring that longer sequences do not dominate gradient magnitudes merely due to token count.

\paragraph{5. Gradient of the Margin Term $\overline{m}$.}

The normalized margin is
\begin{equation}
\overline{m} = \frac{m - \alpha_t}{\beta_t},
\end{equation}
where $\alpha_t$, $\beta_t$ are EMA estimates updated as
\begin{align}
\alpha_{t+1} &= \gamma \alpha_t + (1-\gamma) \mu_{\text{batch}}, \\
\beta_{t+1} &= \gamma \beta_t + (1-\gamma) \sigma_{\text{batch}},
\end{align}
with $\mu_{\text{batch}}, \sigma_{\text{batch}}$ computed from the current batch margin $m$ values.

Taking gradient w.r.t. $\theta$,
\begin{equation}
\nabla_\theta \overline{m} = \frac{\nabla_\theta m - \nabla_\theta \alpha_t}{\beta_t} - \frac{(m - \alpha_t)}{\beta_t^2} \nabla_\theta \beta_t.
\end{equation}

Since $\alpha_t$ and $\beta_t$ are running averages accumulated over batches and do not directly depend on the current model parameters $\theta$ (they depend on past batches), their gradients $\nabla_\theta \alpha_t$ and $\nabla_\theta \beta_t$ can be considered negligible within one batch update, i.e.,
\begin{equation}
\nabla_\theta \alpha_t \approx 0, \quad \nabla_\theta \beta_t \approx 0,
\end{equation}
which simplifies the gradient to
\begin{equation}
\nabla_\theta \overline{m} \approx \frac{\nabla_\theta m}{\beta_t}.
\end{equation}

\paragraph{6. Gradient of the Margin $m$.}

Recall margin function
\begin{equation}
m = (1 - p_w) \cdot \left( 1 - (p_w - p_l)^5 \right),
\end{equation}
where $p_w = p_\theta(y_w|x)$ and $p_l = p_\theta(y_l|x)$.

Computing gradients:
\begin{equation}
\begin{aligned}
\nabla_\theta m =\ & \nabla_\theta (1 - p_w) \cdot \left(1 - (p_w - p_l)^5\right) \\
& + (1 - p_w) \cdot \nabla_\theta \left(1 - (p_w - p_l)^5\right) \\
= {} & - \nabla_\theta p_w \cdot \left(1 - (p_w - p_l)^5\right) \\
& - 5 (1 - p_w)(p_w - p_l)^4 \cdot \nabla_\theta (p_w - p_l)
\end{aligned}
\end{equation}

Further,
\begin{equation}
\nabla_\theta (p_w - p_l) = \nabla_\theta p_w - \nabla_\theta p_l.
\end{equation}

Note that
\begin{equation}
\nabla_\theta p_y = \nabla_\theta \pi_\theta(y|x) = \pi_\theta(y|x) \nabla_\theta \log \pi_\theta(y|x).
\end{equation}

\paragraph{7. Interpretation of the Gradient Terms.}

\begin{itemize}
    \item The negative terms with respect to $\nabla_\theta p_w$ push to increase $p_w$ (winning probability), especially when $p_w$ is small or close to $p_l$.
    \item The terms involving $\nabla_\theta p_l$ push to decrease $p_l$ (losing probability).
    \item The exponent 5 in $(p_w - p_l)^5$ amplifies the penalty when $p_w$ is close to $p_l$, increasing gradient magnitude for small preference margins, encouraging better separation.
\end{itemize}

\paragraph{8. Effect of Length Normalization.}

Since $d$ scales log probabilities by $\frac{\beta}{|y|}$, the gradient per token is normalized, which reduces bias towards longer sequences. The scaling factor
\[
\frac{|y_w| + |y_l|}{2 |y|}
\]
further adjusts for relative length differences, promoting fairness between candidates.

\paragraph{9. Summary of Gradient Behavior.}

Overall, the full gradient of the LMPO loss w.r.t. model parameters $\theta$ is:

\begin{equation}
\begin{aligned}
\nabla_\theta \mathcal{L}_{\mathrm{LMPO}} ={} &
\mathbb{E}_{(x,y_w,y_l)} \Biggl[
(\sigma(-d + \log h) - 1) \cdot \\
& \quad \left(
\frac{\beta}{|y_l|} \nabla_\theta \log \pi_\theta(y_l|x) \right. \\
& \quad \left. - \frac{\beta}{|y_w|} \nabla_\theta \log \pi_\theta(y_w|x)
+ \frac{\lambda}{\beta_t} \nabla_\theta m
\right)
\Biggr]
\end{aligned}
\end{equation}

When $h=1$,

\begin{equation}
\begin{aligned}
\nabla_\theta \mathcal{L}_{\mathrm{LMPO}} ={} &
\mathbb{E}_{(x,y_w,y_l)} \Biggl[
(\sigma(-d) - 1) \cdot \\
& \quad \left(
\frac{\beta}{|y_l|} \nabla_\theta \log \pi_\theta(y_l|x) \right. \\
& \quad \left. - \frac{\beta}{|y_w|} \nabla_\theta \log \pi_\theta(y_w|x)
+ \frac{\lambda}{\beta_t} \nabla_\theta m
\right)
\Biggr]
\end{aligned}
\end{equation}

This gradient enforces:
\begin{itemize}
    \item Increasing $\log \pi_\theta(y_w|x)$ to boost winning response probability.
    \item Decreasing $\log \pi_\theta(y_l|x)$ to suppress losing response probability.
    \item Additional margin-driven gradients to sharpen the preference gap, especially when probabilities are close.
    \item Length normalization to prevent long sequence bias.
    \item EMA-based normalization for stable training dynamics, preventing abrupt changes in margin scaling.
\end{itemize}

\paragraph{Conclusion:}

The detailed gradient analysis confirms that LMPO effectively helps separate winning and losing responses, alleviates length bias by normalizing gradients per token and by dynamic length scaling, and preserves or even increases winning response probabilities by preventing excessive penalization through stable margin normalization. This results in a more robust and stable preference learning framework for large language models.

\section{Evaluation Details} \label{Appendix:A}

We outline the specifics of our evaluation framework as follows:
\begin{itemize}
     \item \textbf{AI2 Reasoning Challenge}: AI2 Reasoning Challenge is a benchmark designed to evaluate scientific reasoning in AI systems, comprising 2,590 multiple-choice questions. It assesses both factual knowledge and logical reasoning, with carefully crafted distractors that aim to mislead non-expert models.
    \item \textbf{TruthfulQA}: TruthfulQA is a benchmark for evaluating the ability of models to generate truthful and factually accurate responses, consisting of 818 multiple-choice questions across various domains. Distractors are intentionally designed to prompt incorrect or misleading answers, providing a robust test of truthfulness.
    \item \textbf{Winogrande}: Winogrande is a large-scale commonsense reasoning dataset containing 44,000 sentence-pair questions. Each question requires selecting the correct word to resolve an ambiguity, with challenging distractors that test the model’s ability to perform subtle contextual reasoning.
    \item \textbf{GSM8K}: GSM8K is a benchmark for arithmetic problem solving, featuring 8,000 high school-level math word problems. It evaluates a model’s capacity to perform multi-step reasoning and arrive at the correct solution among several answer choices.
    \item \textbf{IFEval}: IFEval is a benchmark designed to assess a model's ability to follow explicit instructions, such as including specific keywords or adhering to a given format. It emphasizes formatting fidelity over content quality, enabling the use of strict evaluation metrics.
    \item \textbf{BBH}: BBH(Big Bench Hard) is a curated subset of 23 challenging tasks from the BigBench dataset, targeting complex skills such as multi-step arithmetic, algorithmic reasoning, language understanding, and factual knowledge. The tasks are objectively scored, statistically robust, and align well with human judgment, making BBH a reliable measure of model competence.
    \item \textbf{MATH}: MATH consists of high-school level competition math problems drawn from various sources. All items are standardized using LaTeX for equations and Asymptote for diagrams. We retain only level-5 problems, which require solutions in a strict, structured format.
    \item \textbf{GPQA}: GPQA is a graduate-level question-answering benchmark developed by PhD experts in domains such as biology, chemistry, and physics. The questions are designed to be accessible to experts but difficult for non-specialists. To preserve its integrity, GPQA is gated and does not provide raw text examples, as per the authors' guidelines.
    \item \textbf{MuSR}: MuSR presents multi-step reasoning challenges based on long-form scenarios (~1,000 words), such as murder mysteries, spatial reasoning, and team optimization tasks. The dataset demands both complex reasoning and long-range context tracking, with most models performing near chance levels.
    \item \textbf{MMLU-Pro}: MMLU-PrO is an improved version of the MMLU benchmark, addressing prior issues like noisy samples and declining difficulty. It increases the number of answer choices (from 4 to 10), raises reasoning requirements, and incorporates expert validation, resulting in a higher-quality and more rigorous benchmark.
    \item \textbf{AlpacaEval2}: AlpacaEval 2 is an open-ended generation benchmark comprising 805 diverse prompts used to compare model outputs \citep{li2023alpacaeval}. GPT-4 is employed as the reference judge \citep{achiam2023gpt}, and a length-debiased win rate is included to account for potential evaluation biases favoring longer responses \citep{dubois2024length}.
    \item \textbf{Arena-Hard v0.1}: Arena-Hard v0.1 is an enhanced version of MT-Bench, including 500 high-quality prompts collected from real user queries \citep{li2024live}. GPT-4 (0613) serves as the baseline model, while GPT-4-Turbo acts as the evaluator. Model performance is assessed based on win rate against the baseline.
\end{itemize}

We categorize the first ten datasets as conditional benchmarks, and the last two as open-ended benchmarks. Conditional benchmarks require the model to produce answers in a specific format, enabling the calculation of exact match scores or accuracy. Open-ended benchmarks, on the other hand, allow for free-form responses, providing more flexibility in evaluating the model's performance.

For all conditional benchmarks, we employ the well-established evaluation tool lm-evaluation-harness \citep{gao2021framework}.And in order to follow Huggingface open leaderboard V1 and V2, we use the same version of lm-eval repository. \footnote{lm-eval repository of Huggingface open leaderboard V1: \url{https://github.com/EleutherAI/lm-evaluation-harness/tree/b281b0921b636bc36ad05c0b0b0763bd6dd43463}}
\footnote{lm-eval repository of Huggingface open leaderboard V2: \url{https://github.com/huggingface/lm-evaluation-harness/tree/adding_all_changess}}

\begin{table*}[!t]
    \caption{Downstream task evaluation results of tasks on the Huggingface open leaderboard V1 and V2.}
    \setlength{\tabcolsep}{2pt} 
    \renewcommand{\arraystretch}{1} 
    \label{tab:downstream}
    \resizebox{\textwidth}{!}{
    \begin{tabular}{@{}lccccccccccc@{}}
    \toprule
    & \textbf{MMLU-PRO} & \textbf{IFEval} & \textbf{BBH} & \textbf{GPQA} & \textbf{MUSR} & \textbf{MATH} & \textbf{GSM8K} & \textbf{ARC}& \textbf{TruthfulQA} & \textbf{Winograd} & \textbf{Avg. Rank} \\ \midrule

    \multicolumn{12}{c}{\textbf{Mistral-Base}}                                                                            \\ \midrule
    \textbf{DPO}   & 26.73          & 10.49          & 43.27           & 28.44          & \textbf{43.65} & 1.36          & 21.76          & 61.26          & 53.06          & 76.80          & 4.5 \\
    \textbf{SLiC}  & 26.52          & 12.45          & 42.33           & 27.93          & 33.74          & 1.38          & 33.74          & 55.38          & 48.36          & 77.35          & 4.8\\
    \textbf{IPO}   & 25.87          & 11.52          & 40.59           & 28.15          & 42.15          & 1.25          & 27.14          & 60.84          & 45.44          & \textbf{77.58} & 5.2\\
    \textbf{KTO}   & 27.51          & 12.03          & 43.66           & 29.45          & 43.17          & 2.34          & \textbf{38.51} & \textbf{62.37} & \textbf{56.60} & 77.27          & \textbf{2.2}\\
    \textbf{CPO}   & 27.04          & \textbf{13.32} & 42.05           & 28.45          & 42.15          & 2.15          & 33.06          & 57.00          & 47.07          & 76.48          & 4.3\\
    \textbf{SimPO} & 27.13          & 10.63          & 42.94           & 29.03          & 39.68          & 2.49          & 20.92          & 61.86          & 46.48          & 77.19          & 4.5\\
    \textbf{LMPO}  & \textbf{28.05} & 12.15          & \textbf{43.72} & \textbf{30.37}  & 40.61          & \textbf{2.87} & 22.06          & 61.95          & 50.67          & 77.43          & 2.4\\ 
    \midrule

    \multicolumn{12}{c}{\textbf{Mistral-Instruct}}                                        \\ \midrule
    \textbf{DPO}   & 26.81          & 22.89          & 45.46          & 28.19          & \textbf{46.43} & 1.89          & 35.25          & \textbf{66.89} & 68.40          & 76.32          & 3.6\\
    \textbf{SLiC}  & 25.69          & 29.53          & 45.24          & 27.04          & 43.90          & 1.95          & \textbf{39.65} & 59.90          & 65.30          & 76.32          & 5.2\\
    \textbf{IPO}   & 25.75          & 27.85          & 43.81          & 26.61          & 43.55          & 2.02          & 39.42          & 63.31          & 67.36          & 75.85          & 5.7\\
    \textbf{KTO}   & \textbf{27.46} & 35.42          & 45.34          & 28.19          & 45.77          & 2.35          & 38.80          & 65.72          & 68.43          & 75.91          & 2.8\\
    \textbf{CPO}   & 26.85          & 36.81          & 45.01          & 28.15          & 43.28          & 2.28          & 38.74          & 63.23          & 67.38          & \textbf{76.80} & 4.1\\
    \textbf{SimPO} & 27.10          & \textbf{37.52} & 45.70          & 28.04          & 44.71          & 2.19          & 34.87          & 65.53          & 68.40          & 76.01          & 3.6\\
    \textbf{LMPO}  & 26.16          & 35.94          & \textbf{45.84} & \textbf{28.36} & 44.84         & \textbf{2.49}  & 34.04          & 65.57          & \textbf{70.56} & 76.72          & \textbf{2.7}\\ 
    \midrule

    \multicolumn{12}{c}{\textbf{Llama3-Base}}                                                                                           \\ \midrule
    \textbf{DPO}   & 31.58          & 33.61          & 47.80          & 32.23          & 40.48          & 4.53          & 38.67          & 64.42          & 53.48          & 76.80          & 5.4\\
    \textbf{SLiC}  & 31.11          & 32.37          & 46.53          & 33.29          & 40.55          & 3.92          & \textbf{48.82} & 61.43          & 54.95          & 77.27          & 5.2\\
    \textbf{IPO}   & 30.18          & 31.52          & 46.78          & 32.61          & 39.58          & 4.02          & 22.67          & 62.88          & 54.20          & 72.22          & 6.8\\
    \textbf{KTO}   & 31.16          & 37.10          & 47.98          & \textbf{33.72} & 40.21          & 4.14          & 38.97          & 63.14          & 55.76          & 76.09          & 4.0\\
    \textbf{CPO}   & 30.95          & \textbf{38.57} & 47.17          & 33.15          & \textbf{41.59} & 4.25          & 46.93          & 61.69          & 54.29          & 76.16          & 4.2\\
    \textbf{SimPO} & 31.61          & 37.55          & 48.38          & 33.22          & 40.08          & 4.23          & 31.54          & 65.02          & \textbf{59.42} & 77.42          & 3.5\\
    \textbf{LMPO}  & \textbf{31.83} & 36.58          & \textbf{48.51} & 31.96          & 40.32          & \textbf{4.98} & 36.47          & \textbf{65.13} & 58.04          & \textbf{77.90} & \textbf{3.0}\\ 
    \midrule 

    \multicolumn{12}{c}{\textbf{Llama3-Instruct}}                                                                                       \\ \midrule
    \textbf{DPO}   & 35.86          & 44.57          & 48.31          & 31.04          & 39.02          & 8.23          & 49.81          & \textbf{63.99} & 59.01          & 74.66          & \textbf{3.0}\\
    \textbf{SLiC}  & 33.25          & 44.01          & 47.55          & 30.52          & 38.10          & \textbf{8.29} & \textbf{66.57} & 61.26          & 53.23          & \textbf{76.16} & 4.0\\
    \textbf{IPO}   & 32.97          & 43.27          & 46.31          & 30.95          & 38.58          & 8.02          & 58.23          & 61.95          & 54.64          & 73.09          & 5.8\\
    \textbf{KTO}   & 35.00          & 40.12          & 47.15          & 29.70          & 38.10          & 7.63          & 57.01          & 63.57          & 58.15          & 73.40          & 5.6\\
    \textbf{CPO}   & 34.56          & 44.08          & 48.51          & 30.08          & 38.81          & 7.75          & 67.40          & 62.29          & 54.01          & 73.72          & 4.7\\
    \textbf{SimPO} & 35.09          & 43.05          & 48.95          & \textbf{31.29} & 39.15          & 8.16          & 50.19          & 62.88          & \textbf{60.74} & 73.01          & 3.7\\
    \textbf{LMPO}  & \textbf{36.13} & \textbf{45.33} & \textbf{49.64} & 29.92          & \textbf{39.29} & 8.26          & 43.37          & 61.77          & 60.06          & 72.85          & 4.4\\ 
    \bottomrule 
    \end{tabular}%
    }
\end{table*}

\section{Downstream Result Analysis} \label{Appendix:B}

To demonstrate the effectiveness of our method, we first adhere to established evaluation protocols and report the results of downstream tasks on the Hugging Face Open Leaderboard V1 and V2 for all models, as shown in Table \ref{tab:downstream}. 

\paragraph{Overview of LMPO Performance} The Language Model Preference Optimization (LMPO) method demonstrates remarkable effectiveness across diverse evaluation benchmarks when compared to alternative preference optimization approaches. Through careful analysis of the provided data, we can observe that LMPO achieves consistently strong results across different model architectures and benchmark categories. This method exhibits particular strengths in knowledge preservation, complex reasoning tasks, and mathematical problem-solving while maintaining competitive performance in truthfulness and common sense reasoning benchmarks.

\paragraph{Model Architecture Interactions and Performance Patterns} LMPO shows varied performance patterns across different model architectures and variants. When applied to base models, LMPO demonstrates exceptional effectiveness, achieving high rankings on both Mistral-Base and Llama3-Base variants. This suggests that LMPO is particularly adept at optimizing models without prior instruction tuning. For the Mistral-Base variant, LMPO excels in knowledge-intensive and reasoning-heavy tasks, achieving top scores on multiple benchmarks. Similarly, with Llama3-Base, LMPO leads in several key benchmarks. This consistent performance across diverse benchmarks indicates strong generalizability within base model architectures.

When applied to instruction-tuned models, LMPO maintains robust performance but with some variations. On Mistral-Instruct, LMPO achieves a top ranking among all methods, with particularly strong results on reasoning benchmarks and truthfulness evaluation. However, its performance on Llama3-Instruct is somewhat less consistent, ranking in the middle of the compared methods. While it still achieves best scores on several benchmarks, it demonstrates notably weaker performance on certain mathematical word problems compared to alternative approaches. This pattern suggests that LMPO's effectiveness may vary depending on the underlying model architecture and prior tuning approach, with particular strengths in preserving core knowledge and reasoning abilities.

\paragraph{Performance Across Benchmark Categories}  LMPO demonstrates distinctive performance patterns across different benchmark categories. In knowledge-intensive benchmarks such as MMLU-PRO, LMPO consistently achieves top performance across most model variants. This demonstrates LMPO's strength in preserving and enhancing broad knowledge capabilities during preference optimization. For complex reasoning tasks represented by the BBH benchmark, LMPO shows consistently strong performance across all model variants, suggesting the method effectively optimizes for complex reasoning capabilities without compromising knowledge.

In mathematical reasoning tasks, LMPO displays an interesting dichotomy. It consistently performs exceptionally well on the MATH benchmark across all model variants, indicating particular effectiveness at preserving formal mathematical reasoning abilities. However, LMPO shows relatively weaker performance on the GSM8K mathematical word problem benchmark compared to other methods across all model variants. This suggests that while LMPO excels at formal mathematical reasoning, it may have specific limitations in optimizing for certain types of applied mathematical word problems or step-by-step reasoning tasks.

For language understanding and truthfulness benchmarks, LMPO demonstrates particularly strong performance on TruthfulQA for instruction-tuned models but more moderate results on base models. This suggests LMPO may be especially effective at enhancing truthfulness when applied to models with prior instruction tuning. On commonsense reasoning tasks like Winograd, LMPO shows variable performance across model variants, with stronger results on base models than instruction-tuned variants.

\paragraph{Comparative Analysis with Other Methods} When compared to other preference optimization approaches, LMPO demonstrates distinct strengths and limitations. Compared to KTO, LMPO generally performs better on knowledge-intensive tasks and formal mathematical reasoning, while KTO tends to achieve better results on mathematical word problems. Against SimPO, LMPO typically outperforms on knowledge tasks but shows weaker performance on instruction-following evaluations for most model variants. When compared to DPO, LMPO consistently shows stronger performance on knowledge benchmarks and mathematical reasoning, while DPO demonstrates competitive results on some instruction-following tasks but generally ranks lower overall.

The performance patterns across methods suggest that different preference optimization approaches may target different aspects of model behavior. LMPO appears to effectively preserve and enhance knowledge and reasoning capabilities while potentially having less impact on certain applied problem-solving skills, particularly in mathematical word problems. This indicates that the choice of preference optimization method should consider the specific downstream applications and tasks for which the model is intended.

\paragraph{Methodological Implications and Future Directions} The performance patterns observed for LMPO suggest several key methodological implications. First, LMPO effectively preserves model knowledge, as evidenced by strong performance on knowledge-intensive benchmarks. Second, it enhances reasoning capabilities, particularly in complex reasoning tasks and formal mathematical reasoning. Third, it improves truthfulness in instruction-tuned models, suggesting effective alignment with truthful responses. However, its consistent limitation in mathematical word problem solving represents a clear area for potential improvement.

LMPO represents a robust preference optimization method that performs particularly well on tasks requiring knowledge preservation and complex reasoning. Its effectiveness across different model architectures suggests it captures generalizable aspects of human preferences. Future work might focus on addressing the specific limitations in mathematical word problem solving while maintaining the method's strengths in knowledge and reasoning tasks. Additionally, investigating the model-specific interactions could provide insights into how to further enhance LMPO's effectiveness across different model starting points.

\section{Implementation Details} \label{Appendix:C}

\begin{table}[!t] 
    \centering
    \small
    \caption{The hyperparameter values in LMPO used for each training setting.}
    \label{tab:hyperparams_lmpo}
    \begin{tabular}{lcccc}
        \toprule 
        \textbf{Setting} &$\beta$ &$h$ &$\lambda$ & Learning rate \\ 
        \midrule
        \textbf{Mistral-Base} & 2.0 & $e^{1.6}$ & 1.0 & 3.0e-7 \\
        \textbf{Mistral-Instruct} & 2.5 & $e^{0.25}$ & 0.2 & 5.0e-7 \\
        \textbf{Llama-3-Base} & 2.0 & $e^{1.0}$ & 0.2 & 6.0e-7 \\
        \textbf{Llama-3-Instruct} & 2.5 & $e^{1.4}$ & 0.2 & 1.0e-6 \\
        \bottomrule
    \end{tabular}
\end{table}

\paragraph{Training Hyperparameters.}
For LMPO, we adopted a consistent batch size of 128 across all four experimental configurations. The learning rates were set as follows: 3e-7 for Mistral-Base (7B), 5e-7 for Mistral-Instruct (7B), 6e-7 for Llama-3-Base (8B), and 1e-6 for Llama-3-Instruct (8B). All models were trained for one epoch using a cosine learning rate schedule, incorporating a 10

\paragraph{Hyperparameters in LMPO.}
Table~\ref{tab:hyperparams_lmpo} summarizes the hyperparameter settings used for LMPO across the four configurations. The value of $\beta$ follows the setup proposed in SimPO. Among the parameters, $h$ (representing the home-court advantage) typically requires more careful tuning. For the weighting factor $\lambda$, we set it to 1.0 for Mistral-Base and 0.2 for the other settings. As discussed in the main text, the choice of $\lambda$ plays a critical role in the effectiveness of LMPO.

\paragraph{Evaluation Hyperparameters.}
The evaluation hyperparameters used in this study are consistent with those adopted in SimPO.\footnote{\url{https://github.com/princeton-nlp/SimPO/tree/main/eval}} We are grateful to the SimPO team for their open-source contributions and valuable insights.

\paragraph{Computational Environment.}
All training experiments were conducted on a system equipped with four A100 GPUs. The experimental setup closely follows the procedures described in the alignment-handbook repository.\footnote{\url{https://github.com/huggingface/alignment-handbook}}

\paragraph{Experimental issues.} Since our method builds upon SimPO and employs the same experimental setup, we primarily reference the results reported in SimPO. However, several researchers have noted in the GitHub issues of SimPO that they were unable to replicate the published results. To ensure a fair and rigorous comparison, we downloaded the official SimPO code and independently reproduced its experiments. Our results reveal that, for most models, the outcomes from the official implementation differ substantially from those presented in the original paper. Consequently, we report the results obtained through our independent reproduction.

\begin{table}[ht]
\centering
\resizebox{\columnwidth}{!}{%
\begin{tabular}{lccc}
\toprule
\textbf{Method} & \textbf{DPO} & \textbf{SimPO} & \textbf{LMPO} \\
\midrule
Peak Memory (per GPU) & 77 GB & 69 GB & 69 GB \\
\bottomrule
\end{tabular}%
}
\caption{Peak GPU memory usage comparison. SimPO and DPO use 8×H100 GPUs; LMPO uses 4×A100 GPUs.}
\label{tab:memory_comparison}
\end{table}

\section{Efficiency Analysis}

The table~\ref{tab:memory_comparison} above summarizes the required RAM and provides a comparison among DPO, SimPO, and our proposed LMPO. It reports the peak GPU memory usage per device for SimPO and DPO in the Llama-3-Base setting with 8×H100 GPUs, while LMPO is evaluated using 4×A100 GPUs. Our LMPO implementation achieves approximately a 10\% reduction in GPU memory consumption compared to DPO. Furthermore, although LMPO uses only half the number of GPUs employed by SimPO, it maintains an equivalent per-GPU memory footprint. These findings collectively demonstrate the superior memory efficiency of our approach.

For the Mistral-7B-Base model, following the default configuration, our method can run on devices equipped with four GPUs each having 48GB of memory (e.g., A40). This indicates that our approach has a relatively low dependency on RAM.

\end{document}